\documentclass{bmvc2k}

\title{BioFaceNet: Deep Biophysical Face \\ Image Interpretation}

\addauthor{Sarah Alotaibi \textsuperscript{1,2}}{https://www.cs.york.ac.uk/cvpr/member/sarah}{1}
\addauthor{William A.~P.~Smith \textsuperscript{1}}{https://www-users.cs.york.ac.uk/wsmith}{1}

\addinstitution{
	\textsuperscript{1}Department of Computer Science\\
	University of York\\
	York, UK \\
	\vspace{0.1mm}
	\textsuperscript{2}Computer Science Department\\
	King Saud University\\
	Riyadh, KSA
}

\runninghead{Alotaibi and Smith}{BioFaceNet: Deep Biophysical Face Interpretation}

\def\etal{\emph{et al}\bmvaOneDot}
\usepackage{times}
\usepackage{epsfig}
\usepackage{graphicx}
\usepackage{amsmath}
\usepackage{amssymb}
\usepackage{amssymb,bm}
\newcommand{\R}{\mathbb{R}}

\usepackage[export]{adjustbox}
\usepackage{wrapfig}
\pdfoutput=1
\begin{document}

\maketitle
\begin{abstract}
In this paper we present BioFaceNet, a deep CNN that learns to decompose a single face image into biophysical parameters maps, diffuse and specular shading maps as well as estimating the spectral power distribution of the scene illuminant and the spectral sensitivity of the camera. The network comprises a fully convolutional encoder for estimating the spatial maps with a fully connected branch for estimating the vector quantities. The network is trained using a self-supervised appearance loss computed via a model-based decoder. The task is highly underconstrained so we impose a number of model-based priors. Skin spectral reflectance is restricted to a biophysical model, we impose a statistical prior on camera spectral sensitivities, a physical constraint on illumination spectra, a sparsity prior on specular reflections and direct supervision on diffuse shading using a rough shape proxy. We show convincing qualitative results on in-the-wild data and introduce a benchmark for quantitative evaluation on this new task.
\end{abstract}
\section{Introduction}
\label{sec:intro}

Providing a physical explanation of the appearance of a face is a longstanding goal in computer vision. From 3D face capture in computer graphics to extracting identity specific information for face recognition, there are clear benefits to being able to separate intrinsic properties of the face from extrinsic scene conditions when the image was captured. It is therefore surprising that the vast majority of methods that study face appearance use generic models that are applicable to any object and do not take into account constraints provided by the specific appearance of a face. For example, even in state-of-the-art deep learning based methods it is often assumed that faces are Lambertian diffuse reflectors \cite{kim17InverseFaceNet,Shu_2017_CVPR,SfSNet} and they ignore the specular component (resulting from oily skin or sweat) and subsurface effects. Where diffuse albedo (i.e.~intrinsic colour of the skin) is explicitly modelled this is usually done with a statistical model \cite{Tewari_2017_ICCV,tewari18FaceModel,nhan2015beyond,Li_2014,blanz1999morphable} or, in the case of intrinsic image decomposition approaches, with an unconstrained albedo map \cite{Shu_2017_CVPR,SfSNet}. However, it is known that skin colour forms a curved manifold in RGB space \cite{Claridge2003489,preece2003imaging,preece2004spectral} spanned by the main pigments in skin. Models that do not impose this biophysical constraint can generate implausible skin colours and linear models will require redundant dimensions to capture the nonlinear subspace. Besides providing a strong constraint on plausible skin colours, modelling in the biophysical domain also has advantages from an application point of view as it allows intuitive editing of parameter maps with physical meaning. These advantages motivate our decision to model face appearance using a biophysical model.

\begin{figure}[!t]
\centering
\includegraphics[trim={0cm 0.1cm 0cm 0cm},width=\textwidth,clip=true]{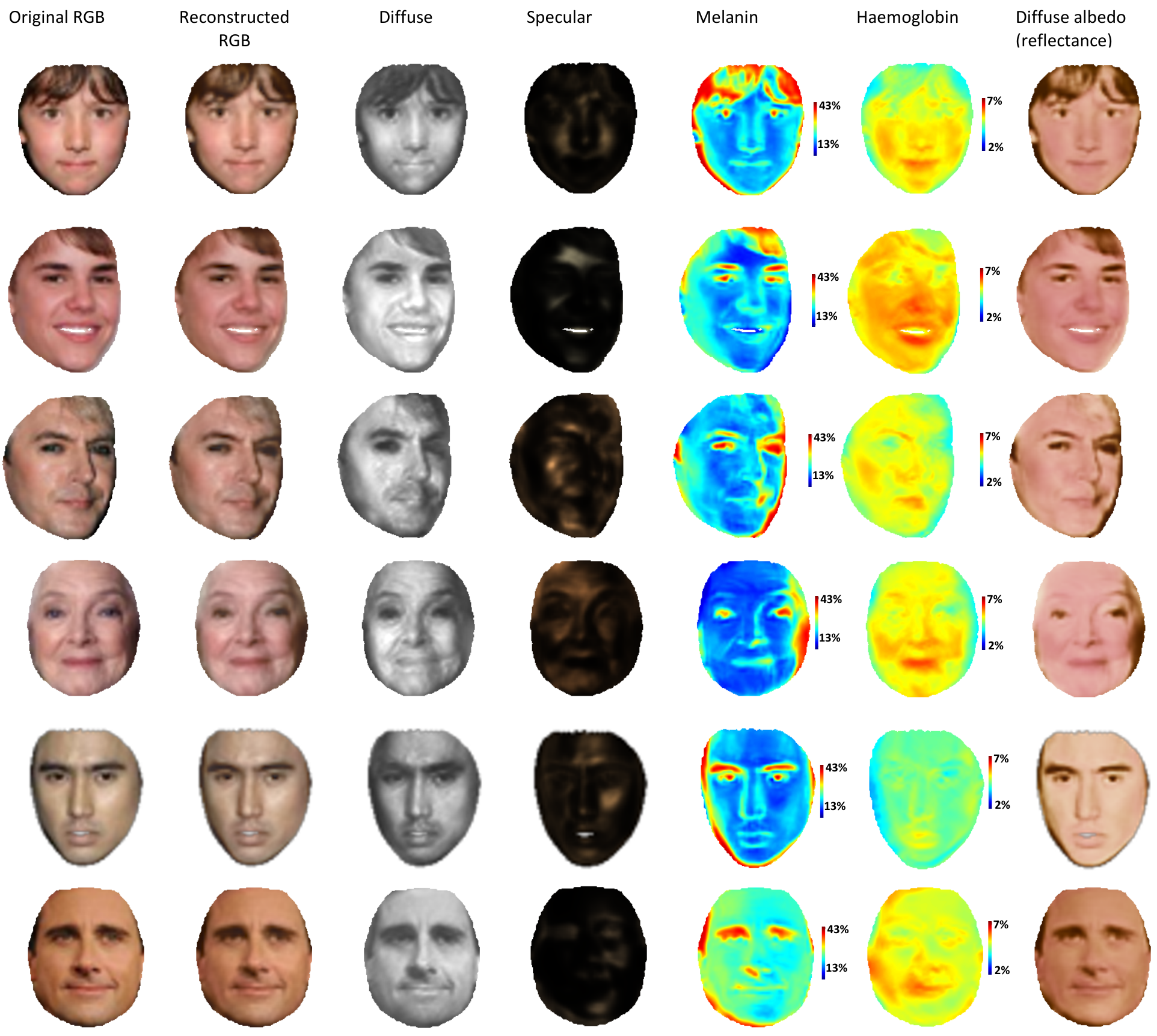}
\caption{Sample decomposition results on CelebA dataset \cite{liu2015deep}. Col.~1: input, col.~2: reconstruction  (gamma correction has been applied on input and output for visualisation). Col.~3: diffuse shading $i_d$. Col.~4: specular shading $i_s$. Col.~5-6: melanin $f_{\textrm{fmel}}$ and haemoglobin $f_{\textrm{fblood}}$. Col.~7: diffuse albedo from biophysical maps.}
\label{fig:decomposition}
\end{figure}

In this paper, we propose BioFaceNet: a deep convolutional neural network that learns to decompose a single RGB image into intrinsic components in the spectral domain. This is an ill-posed problem and so careful modelling and constraint is required to render the problem tractable. This knowledge is encapsulated in a model-based decoder that is used to train a CNN-based encoder. We combine a dichromatic reflectance model and biophysical spectral skin colouration model in order to decompose face appearance into specular and diffuse shading and distribution maps for two biophysical parameters (melanin and haemoglobin). In addition we estimate spectral illumination and camera sensitivity, constrained by physical and statistical models respectively. See \ref{fig:decomposition} for some sample results.

\subsection{Deep face appearance decomposition} 
In recent years, deep neural networks have been applied to estimate face parameters such as geometry, appearance properties and the results obtained are inspiring. Recent studies on face reconstruction \cite{Tewari_2017_ICCV,Shu_2017_CVPR,SfSNet} rely on statistical face models to constrain geometry and appearance estimation. Tewari \etal \cite{Tewari_2017_ICCV} introduce MoFA: a self-supervised learning approach to train a model-based autoencoder CNN architecture. The CNN is able to fit a 3D morphable model \cite{blanz1999morphable} to single images by estimating shape and reflectance and regress scene illumination. The encoder learns to extract face parameters and the decoder uses a differentiable image formation to construct an image that allows unsupervised training on real images. The self-supervised loss is the error between the constructed image and the input. 
Kim \etal \cite{kim17InverseFaceNet} introduced InverseFaceNet that estimate 3DMM parameters including colour reflectance and illumination. The CNN was trained on synthetic dataset and used a breeding method to increase variability in training dataset. Again the skin reflectance estimated based on statistical appearance model \cite{blanz1999morphable}. 
Shu \etal \cite{Shu_2017_CVPR} proposed unsupervised autoencoder networks to learn facial appearance's components: albedo, normal, and lighting. They combined constraints on each components with an adversarial loss on on image reconstruction. Their decomposition is still not realistic.   
On the other hand, Sengupta \etal \cite{SfSNet} start with supervised training on synthetic data and later finetune this network on real data to achieve: albedo, normal, and lighting estimates and these components are used later on pseudo-supervision stage. The image formation relies on Lambertian reflectance. A photometric reconstruction loss is applied to validate the composition.

\subsection{Biophysical skin modelling}
Modelling the appearance of human skin is fundamentally challenging, due to the complexity of its layered-structure and its optical properties. 
Tsumura \etal \cite{Tsumura:2003:ISC} presented an image-based method to recover the concentrations of melanin and haemoglobin from colour face image using Independent Component Analysis (ICA). Their method is restricted to specific light and camera combinations. Donner \etal \cite{donner2008layered} use multispectral and polarised light to derive biophysical skin parameter maps from a 2D planar sample. Gitlina \etal \cite{gitlina2018practical} have combined polarised spherical gradient illumination with multispectral lighting to acquire spectral skin reflectance.
Other studies focused on building models that accurately simulate face appearance by applying skin optics with biophysical components to reproduce spectral and spatial responses. Krishnaswamy and Baranoski \cite{krishnaswamy2004biophysically} introduced the BioSpec model, with about twenty-four physically-meaningful parameters to simulate the light interaction within five layers of human skin. Biospec is computationally expensive, and very difficult to invert. Claridge and co-authors  \cite{Claridge2003489,preece2003imaging,preece2004spectral}  combined a calibrated camera with a two or three parameter model based on Kubelka-Munk theory to measure skin parameters. Jimenez \etal \cite{JIMENEZ2010} presented skin model to simulate dynamic effects caused by facial expressions.


\section{Preliminaries}\label{sec:prelim}

Our model-based decoder simulates the spectral formation of an RGB image. This requires a number of basic components that we describe in this section. A number of assumptions underlie the choice of these components. We assume: 1.~Images are captured by a camera that correctly white balances the scene and uses a fixed gamma, 2.~Scene illumination is spectrally uniform, 3.~Skin reflectance follows the dichromatic reflectance model. Assumption 2 is clearly violated in real images. For example, shadowed regions will be illuminated by different spectra to directly lit parts of the face. However, allowing spatially varying illumination spectra adds significant complexity and ambiguity to the problem and we leave this to future work.

\subsection{Spectral image formation}

A tristimulus RGB image arises from an integration over wavelength, $\lambda$, of the product of scene radiance $L(\lambda)=E(\lambda)R(\lambda)$ (itself the product of illumination and reflectance spectra) and camera spectral sensitivity:
\begin{equation}
    i_c = \int_0^{\infty} E(\lambda)R(\lambda)S_c(\lambda)d\lambda,
\end{equation}
where $E$ is the spectral power distribution (SPD) of the illuminant, $R$ the spectral reflectance of the surface and $S_c$ the spectral sensitivity of the camera in colour channel $c\in\{R,G,B\}$.

\subsection{Wavelength-discrete spectral image formation}

We approximate this continuous model by discretising wavelength at $D$ locations:
\begin{equation}
    \mathbf{i}_{\textrm{raw}} = [i_R, i_G, i_B]^T = \mathbf{S}^T\textrm{diag}(\mathbf{e})\mathbf{r},\label{eqn:wavdisim}
\end{equation}
where $\mathbf{S}\in\R^{D\times 3}$, $\mathbf{e}\in\R^D$ and $\mathbf{r}\in\R^D$ are the wavelength-discrete versions of the camera sensitivities, illuminant SPD and spectral reflectance respectively. We use $\mathbf{s}_c\in\R^D$ to refer to the column of $\mathbf{S}$ corresponding to colour channel $c$.

\subsection{Colour transformation pipeline}

The raw colours measured by a sensor, $\mathbf{i}_{\textrm{raw}}$, are transformed by the camera in order to produce perceptually pleasing images. The purpose is to normalise for lighting and sensor specific effects and apply a nonlinear mapping to compress intensities to a dynamic range that can be stored and displayed. The precise details of this pipeline are camera-specific however we assume the following generic model that is a good approximation for most cameras:
\begin{equation}
    \mathbf{i}_{\textrm{linRGB}} = \mathbf{T}_{\textrm{xyz2rgb}}\mathbf{T}_{\textrm{raw2xyz}}(\mathbf{S})\mathbf{T}_{\textrm{wb}}(\mathbf{S},\mathbf{e})\mathbf{i}_{\textrm{raw}}.
    \label{eqn:colourpipeline}
\end{equation}
The first transformation, $\mathbf{T}_{\textrm{wb}}(\mathbf{S},\mathbf{e})\in\R^{3\times 3}$, performs white balancing for a given illuminant and camera. Specifically, it divides each channel by the colour of the light source as recorded by the sensor:
\begin{equation}
    \mathbf{T}_{\textrm{wb}}(\mathbf{S},\mathbf{e}) = \textrm{diag}(\mathbf{S}^T\mathbf{e})^{-1}.
\end{equation}
The second transformation, $\mathbf{T}_{\textrm{raw2xyz}}(\mathbf{S})\in\R^{3\times 3}$, converts from the camera-specific colour space to the standardised XYZ space:
\begin{equation}
    \mathbf{T}_{\textrm{raw2xyz}}(\mathbf{S}) = \mathbf{CS}^+,
\end{equation}
where $\mathbf{C}\in\R^{D\times 3}$ contains the wavelength discrete CIE-1931 2-degree color matching function and $\mathbf{S}^+$ is the pseudoinverse of $\mathbf{S}$ \cite{jiang2013space}. This is a least squares solution to transform the camera's spectral sensitivities to the CIE standard. We additionally rescale each row such that its sum is unity to preserve white balance such that $\mathbf{T}_{\textrm{raw2xyz}}(\mathbf{S})\mathbf{1}=\mathbf{1}$. The final transformation is a fixed matrix to convert to sRGB space:
\begin{equation}
    \mathbf{T}_{\textrm{xyz2rgb}}=\begin{bmatrix}
3.2406 & -1.537 & -0.498 \\
-0.968 & 1.8758 & 0.0415 \\
0.0557 & -0.204 &  1.0570\\
\end{bmatrix}
\end{equation}
after which a final nonlinear gamma transformation is applied:
\begin{equation}
    \mathbf{i}_{\textrm{sRGB}} = (1+a)\mathbf{i}_{\textrm{linRGB}}^{1/\gamma}-a,
\label{eqn:wholePipeline}
\end{equation}
where we assume $a=0.055$ and $\gamma=2.4$.

\subsection{Multispectral dichromatic model}

The dichromatic model \cite{shafer1985using} assumes that scene radiance, $L(\lambda)$, is a sum of body (diffuse) and surface (specular) reflected components. Further, it divides each source of radiance into a part that depends on geometry (informally ``shading'') and a wavelength dependent part (informally ``colour''). The body reflection arises from subsurface scattering and modifies the SPD of the light through absorption whereas the surface reflectance happens at the interface and does not, meaning the model can be written as:
\begin{equation}
    L(\lambda) = E(\lambda)(i_dR(\lambda) + i_s),
\end{equation}
where $i_d\in\R_{\geq 0}$ and $i_s\in\R_{\geq 0}$ are the diffuse and specular shading respectively. In wavelength-discrete terms, this becomes:
\begin{equation}
    \mathbf{l} = \textrm{diag}(\mathbf{e})(i_d\mathbf{r} + i_s). \label{eqn:discretemodel}
\end{equation}
Combining \eqref{eqn:wavdisim} and \eqref{eqn:discretemodel} provides our appearance model.

\subsection{Statistical camera model}

The space of camera spectral sensitivities has been shown to be low dimensional. Using PCA to build a statistical model, Jiang \etal \cite{jiang2013space} showed that two dimensions were sufficient to capture 97\% of the variance of a data set of 28 empirically measured sensitivities. Accordingly, any spectral sensitivity can be approximated as:
\begin{equation}
    \textrm{vec}\left(\mathbf{S}(\mathbf{b})\right) = \mathbf{P}\textrm{diag}(\sigma_1,\dots,\sigma_N)\mathbf{b}+\textrm{vec}(\bar{\mathbf{S}}),
\end{equation}
where $\mathbf{P}\in\R^{3D\times N}$ contains the first $N$ principal components, $\sigma^2_1,\dots,\sigma^2_N$ are the corresponding eigenvalues, $\bar{\mathbf{S}}\in\R^{3D}$ is the mean sensitivity and $\mathbf{b}\in\R^N$ is the parametric representation of $\mathbf{S}$. We use $N=2$ dimensions. Under the assumption that the original data is Gaussian distributed then the parameters are normally distributed: $\mathbf{b}\sim\cal{N}(\mathbf{0},\mathbf{I})$.

\subsection{Physical lighting model}
Our spectral illumination model is physically-based. We assume that the scene illumination can be approximated by a linear combination of CIE standard illuminants A, D and F respectively representing incandescent light, phases of daylight and fluorescent lights of various composition. Illuminant D requires an additional parameter representing the colour temperature ranging from 4,000 to 25,000K. Illuminant F is itself a linear combination of 12 measured fluorescent sources. Hence, our illumination model is given by:
\begin{equation}
    \mathbf{e}(w_{A},w_{D},t,w_{F1},\dots,w_{F12}) = w_{A}\mathbf{e}_A + w_{D}\mathbf{e}_D(t) + w_{F1}\mathbf{e}_{F1} + \dots + w_{F12}\mathbf{e}_{F12},\label{eqn:lightmodel}
\end{equation}
where $w_{A},w_{D},w_{F1},\dots,w_{F12}\in\R_{\geq 0}$ are the weights for each illuminant type, $t$ is the correlated color temperature and $\mathbf{e}_A,\mathbf{e}_D(t),\mathbf{e}_{F1},\dots,\mathbf{e}_{F12}\in\R^D$ are the spectra of the standard illuminants.

\section{Biophysical spectral skin model}
\label{label:BioSkin}
We now constrain the multispectral dichromatic model in \eqref{eqn:discretemodel} using a biophysical human skin model. This has only two free biophysical parameters, such that the resulting biophysical dichromatic model has four unknowns per pixel in total. Our biophysical spectral reflectance model is closely related to a number of existing models \cite{Claridge2003489,JIMENEZ2010,krishnaswamy2004biophysically,preece2004spectral,donner2008layered}. However, for the challenging task we seek to solve, we focus on simplicity and the minimum number of free parameters. Specifically: the melanin and haemoglobin concentration that vary spatially while all other parameters are based on validated approximation functions or measured data for healthy skin \cite{jacques1998skin,Alotaibi_2017_ICCV,ANDERSON198113,THODY1991340,JIMENEZ2010,prahl1999optical,Flewelling1999,krishnaswamy2004biophysically}.  
We used a simplified two layered skin structure model of \cite{alotaibi2019decomposing}. The epidermis is the outer layer containing the melanin pigment, originated from melanosomes cells, that absorbs the blue wavelengths and the rest of the light is mainly forward scattered. The deeper layer is the dermis containing blood vessels that carry the haemoglobin pigment and absorbs light in the blue and green wavelengths while the rest of the light is reflected back and reaches the epidermis where again absorption and forward scattering occur before light exits skin. This simplified model is written as:
\begin{equation}
R(f_{\textrm{fmel}}, f_{\textrm{fblood}},\lambda)= T_{\textrm{epidermis}}(f_{\textrm{mel}},\lambda) ^{2} R_{\textrm{dermis}}(f_{\textrm{fblood}},\lambda).    
\end{equation}
where $f_{\textrm{fmel}}$ is the epidermal melanosomes volume fraction and falls in the range $f_{\textrm{fmel}}^{\textrm{min}}=1.3\%\dots f_{\textrm{fmel}}^{\textrm{max}}=43\%$, $f_{\textrm{fblood}}$ is the dermal blood volume fraction and falls in the range $f_{\textrm{fblood}}^{\textrm{min}}=2\%\dots f_{\textrm{fblood}}^{\textrm{max}}=7\%$, $T_{\textrm{epidermis}}(f_{\textrm{fmel}},\lambda)\in [0,1]$ is the proportion light transmitted through the epidermis (twice) and is modelled using the Lambert-Beer law, $R_{\textrm{dermis}}(f_{\textrm{fblood}},\lambda)\in [0,1]$ is the proportion of light reflected from the dermis and is modelled by Kubelka-Munk theory. 
In wavelength-discrete terms, we write $\mathbf{r}(f_{\text{fmel}},f_{\text{fblood}})\in\mathbb{R}^D$ as the vector of diffuse spectral reflectance which can be substituted into \eqref{eqn:discretemodel}.

\section{Architecture}

\begin{figure}[t!]
    \centering
    \includegraphics[width=0.9\textwidth,clip=true, trim={0cm .1cm 0cm 0cm}]{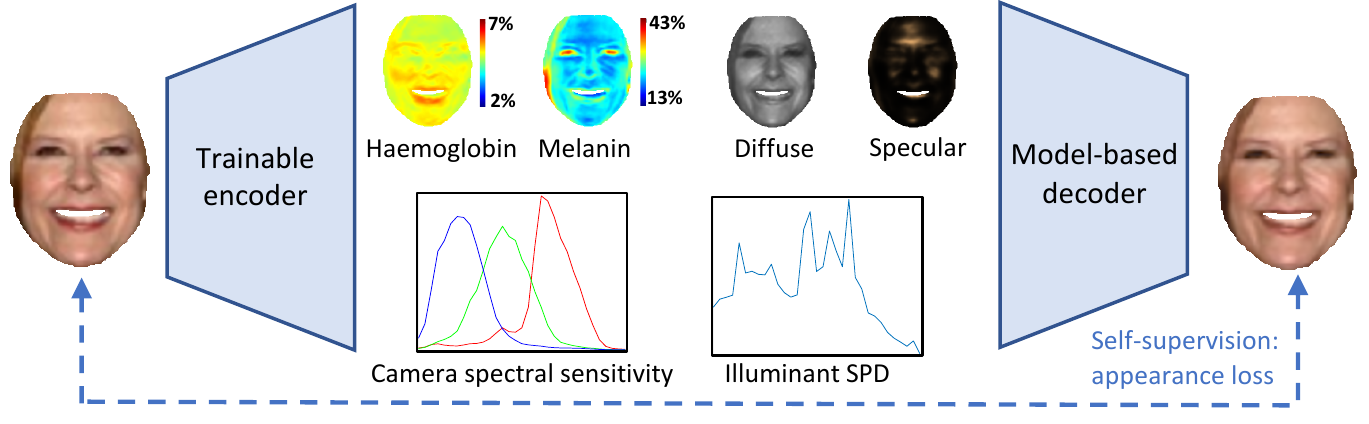}
    \caption{Overview of BioFaceNet. From an RGB image we estimate haemoglobin, melanin, diffuse and specular maps, camera spectral sensitivity and illumination. Self-supervision is provided by a model-based decoder that renders an image.}
    \label{fig:overview}
\end{figure}

Our overall architecture is shown in Fig.~\ref{fig:overview}. At the most abstract level, this consists of a trainable convolutional encoder that estimates semantically meaningful parameters and a fixed, differentiable, model-based decoder that implements spectral image formation to transform these parameters back into an image. The semantic representation consists of four image quantities (the two biophysical parameter maps and diffuse and specular shading maps) and two vector quantities (parameters for the physical lighting and statistical camera models).

\subsection{Trainable encoder}
The encoder is a CNN and itself has an encoder/decoder architecture, all of which is trainable. We invert the nonlinear gamma on the input image such that the input is in linear space and appearance losses are calculated without applying gamma, i.e.~also in linear space. We found that this gave more stable convergence than using nonlinear input images and applying gamma to our rendered output. The maps are predicted by a fully convolutional network with skip connections, following a U-net \cite{ronneberger9351u} style architecture but with separate decoders for each map. The encoder/decoder consists of three convolutions per resolution with filter sets: $32 \ast 3 \times 3$, $64 \ast 3 \times 3$, $128 \ast 3 \times 3$, $256 \ast 3 \times 3$ and $512 \ast 3 \times 3$. Each convolution is followed by batch normalisation, ReLU nonlinearity and finally max-pooling. From the lowest spatial resolution, a fully connected branch predicts the vector quantities $\theta=(\mathbf{b},w_{A},w_{D},t,w_{F1},\dots,w_{F12})$. Since the encoder used to predict all 6 quantities is shared, this helps the encoder learn to disentangle the interaction of the different quantities.

Since the estimated quantities have physical meaning, they are bounded or subject to positivity constraints. The diffuse and specular maps must be positive so the raw estimates are exponentiated. The haemoglobin and melanin maps are bounded by the physically-plausible ranges in Sec.~\ref{label:BioSkin} which we map to the range $[-1,1]$. Hence, the raw estimates are passed through a sigmoid function, scaled by 2 and shifted by $-1$. Similarly, the camera parameters, $\mathbf{b}$, are transformed to the range $[-3,3]$ (assuming $\pm 3$ standard deviations from the mean captures sufficient variation) and the correlated colour temperature, $t$, is transformed to the range $[1,22]$.
There is an intrinsic scale ambiguity between the overall intensity of the light source and the diffuse/specular shading (i.e.~the same image can be obtained by multiplying the illumination by 2 and dividing the shading maps by 2). We resolve this by rescaling all standard illuminants to have unit sum, $\mathbf{e}^T\mathbf{1}=1$, and then taking only \emph{convex} combinations in \eqref{eqn:lightmodel}, i.e.~we enforce that $w_A+w_D+w_{F1} + \dots + w_{F12}=1$. This is achieved by passing the weights predicted by the encoder through a softmax layer which also ensures their positivity. This guarantees $\mathbf{e}(w_{A},w_{D},t,w_{F1},\dots,w_{F12})^T\mathbf{1}=1$ and so fixes the scale of illumination.

\begin{figure}[t!]
 \centering
    \includegraphics[trim={0cm 0.2cm 0cm 0cm},width=0.83\textwidth]{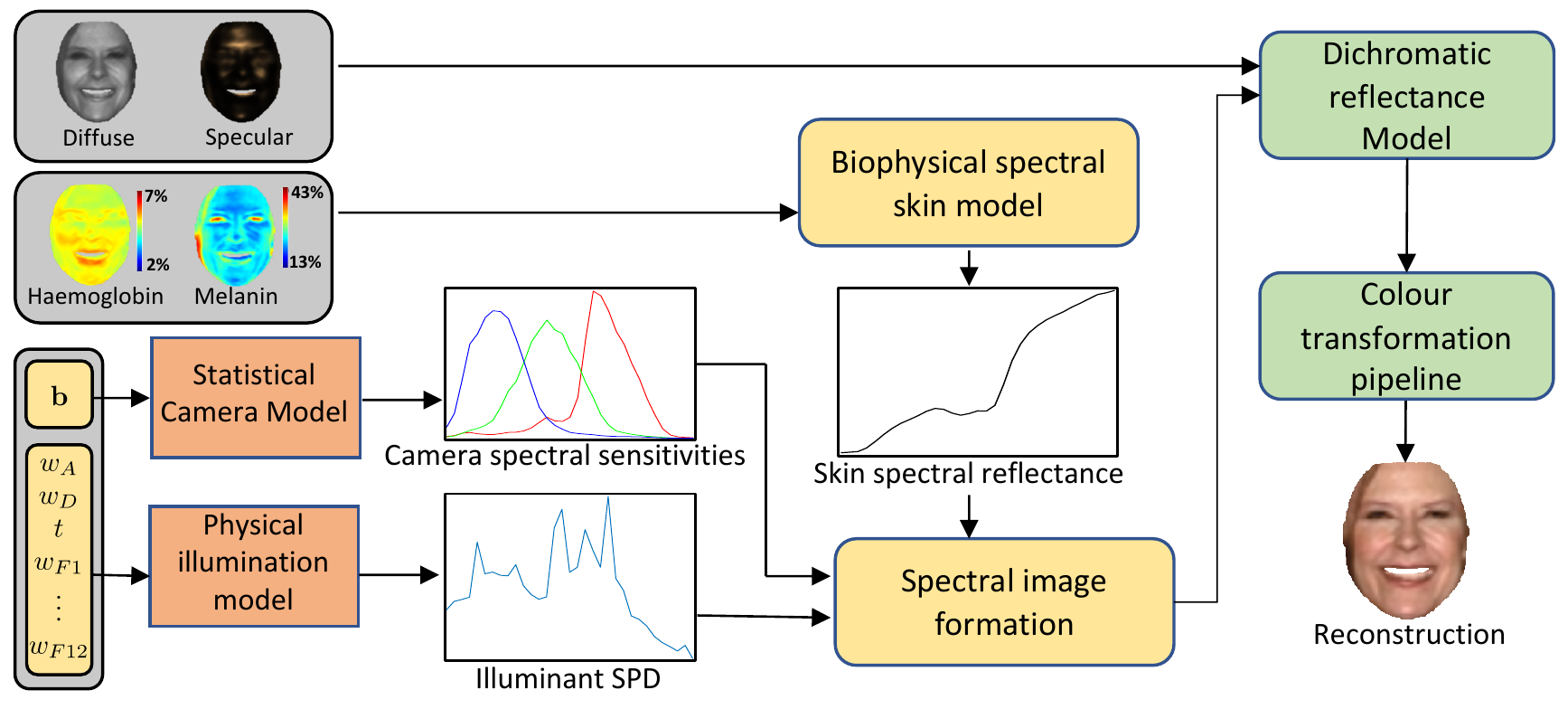}
    \caption{Components of our model-based decoder. Grey boxed items represent the output of the trainable encoder.}
    \label{fig:Model-basedDecoder}
\end{figure}

\subsection{Model-based decoder}

The model-based decoder implements the components described in Sections \ref{sec:prelim} and \ref{label:BioSkin} as shown in Fig.~\ref{fig:Model-basedDecoder}. All components are implemented in a differentiable manner, such that the gradients of the subsequent loss functions can be backpropagated through the decoder and into the trainable encoder. For efficiency, we precompute skin spectral reflectance at discrete values of the biophysical parameters within their plausible ranges and store in a 2D look up table. We then use differentiable bilinear interpolation to compute reflectance for continuous parameter values. In the colour transformation pipeline, computing $\mathbf{T}_{\textrm{raw2xyz}}$ requires taking a pseudoinverse of the camera spectral sensitivity. While this can be done in-network, for efficiency and stability we precompute $\mathbf{T}_{\textrm{raw2xyz}}$ as a lookup table as a function of $\mathbf{b}$ and again use bilinear interpolation.

\subsection{Losses}
We train our network to minimise four losses:
\begin{equation}
{\mathcal{L}} = w_1\mathcal{L}_\mathrm{appearance} + w_2\mathcal{L}_\mathrm{CameraPrior} + w_3\mathcal{L}_\mathrm{SpecSparsity} + w_4\mathcal{L}_\mathrm{ShadingSup}
\end{equation}
The first is a self-supervised appearance loss measuring the difference between the input $\mathbf{i}_{\textrm{linObs}}$ and reconstructed images (see Fig.~\ref{fig:overview}): ${\mathcal{L}_\mathrm{appearance}} = \left\| \mathbf{i}_{\textrm{linRGB}}  - \mathbf{i}_{\textrm{linObs}} \right\|_2^2$. Using this loss alone allows the network to converge to trivial solutions with physically meaningless decomposition of appearance. To constrain the problem we introduce three additional priors. We enforce a statistical prior loss on the camera sensitivity parameters: $\mathcal{L}_\mathrm{CameraPrior}=\|\mathbf{b}\|_2^2$. Assuming lighting is sparse and the face surface smooth, we can assume that specular reflections are sparse and so impose an L1 sparsity prior on the specular shading $\mathcal{L}_\mathrm{SpecSparsity}=\|\mathbf{i}_s\|_1$. Finally, we provide some weak direct supervision of the diffuse shading. Following \cite{Shu_2017_CVPR}, we use an approximate normal map and spherical harmonic parameters obtained by a rough fit of a 3D morphable model and use this to compute pseudo ground truth diffuse shading, $i_d^{\text{PGT}}$. There is an unknown scale ambguity between this shading and the one estimated by our network. So we compute the optimal scale, $s$, using simple linear regression without the intercept term and apply this to our estimate before computing an L2 shading loss: $\mathcal{L}_\mathrm{ShadingSup}=\|si_d - i_d^{\text{PGT}}\|_2^2$. The three pixel-wise losses are summed over pixels and normalised by the number of foreground masked pixels.

\section{Experiments}

\begin{figure}
\centering
\includegraphics[trim={0cm 0.2cm 0cm 0cm},width=7.1cm]{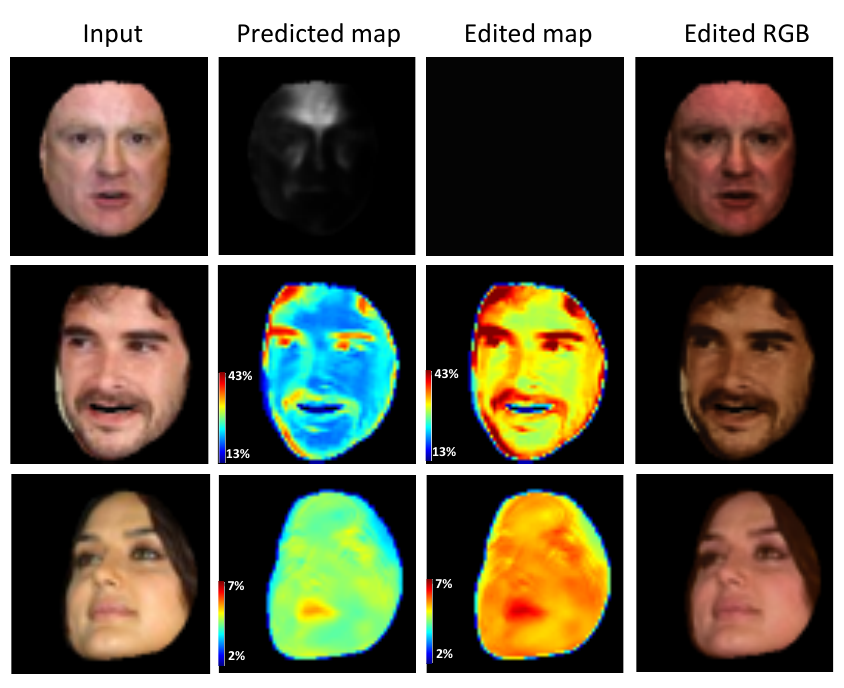}
\caption{Results of editing specular, melanin and haemoglobin maps respectively.}
\label{fig:CNNApplications}
\end{figure}

\begin{table}[!t]
\resizebox{\columnwidth}{!}{
 \begin{tabular}{|c|c|c|c|c|c|c|} 
 \hline
 Algorithm & Diffuse & Specular & Albedo & Melanin & Haemoglobin & Reconstruction\\ [0.5ex] 
 \hline\hline
 \textbf{SfSnet} & 0.1898 & N/A & 0.1452 & N/A & N/A &  0.2831\\ 
 \hline
 \textbf{BioFaceNet} & \textbf{0.1720} & 0.0875 & \textbf{0.1154} & 0.5972 & 0.2912 & \textbf{0.0713}\\ [1ex] 
 \hline
\end{tabular}
}
 \caption{The quantitative result using the root mean square error (RMSE)}
 \label{quantitative1}
\end{table}	

We implement our network using the autonn wrapper for MatConvNet. We train on a 50k subset of the CelebA dataset \cite{liu2015deep} as in \cite{Shu_2017_CVPR} using SGD, a learning rate of $1\textrm{e}-5$ and set the loss weights to $(w_1,w_2,w_3,w_4)=(1\textrm{e}-3,1\textrm{e}-4,1\textrm{e}-5,1\textrm{e}-5)$.

In Fig.~\ref{fig:decomposition}, we present qualitative results on unseen test images from CelebA. It is clear that the lips and flushed cheeks appear in the haemoglobin maps with high concentrations and the overall melanin maps reflects skin colour accurately. The specular maps detect the specular reflections and the diffuse shading is blurred as a result of subsurface scattering. We compute the diffuse albedo directly from the biophysical spectral reflectance. The fourth row shows a failure case where shadowing is interpreted as high haemoglobin.
In Fig.~\ref{fig:CNNApplications}, we show results of an editing application. We edit an estimated map, then we recompute the final image as in \eqref{eqn:wholePipeline}.
In the first row, we remove specular reflections by setting the specular to constant map. The apparent changing colours between these images after reove the specular is consistent with \cite{alotaibi2019decomposing} where the multispectra data is used. In the second row, we increase the melanin pigment by 0.6 and this shows a darker skin of the face appearance such as the face has been sun-tanned. In the last row, we scale the haemoglobin by 0.5 and this gives a flushed appearance such as if the face is overheated. 

\begin{figure}[!t]
\centering
\includegraphics[trim={0cm 0.2cm 0cm 0cm},width=0.8\textwidth,clip=true]{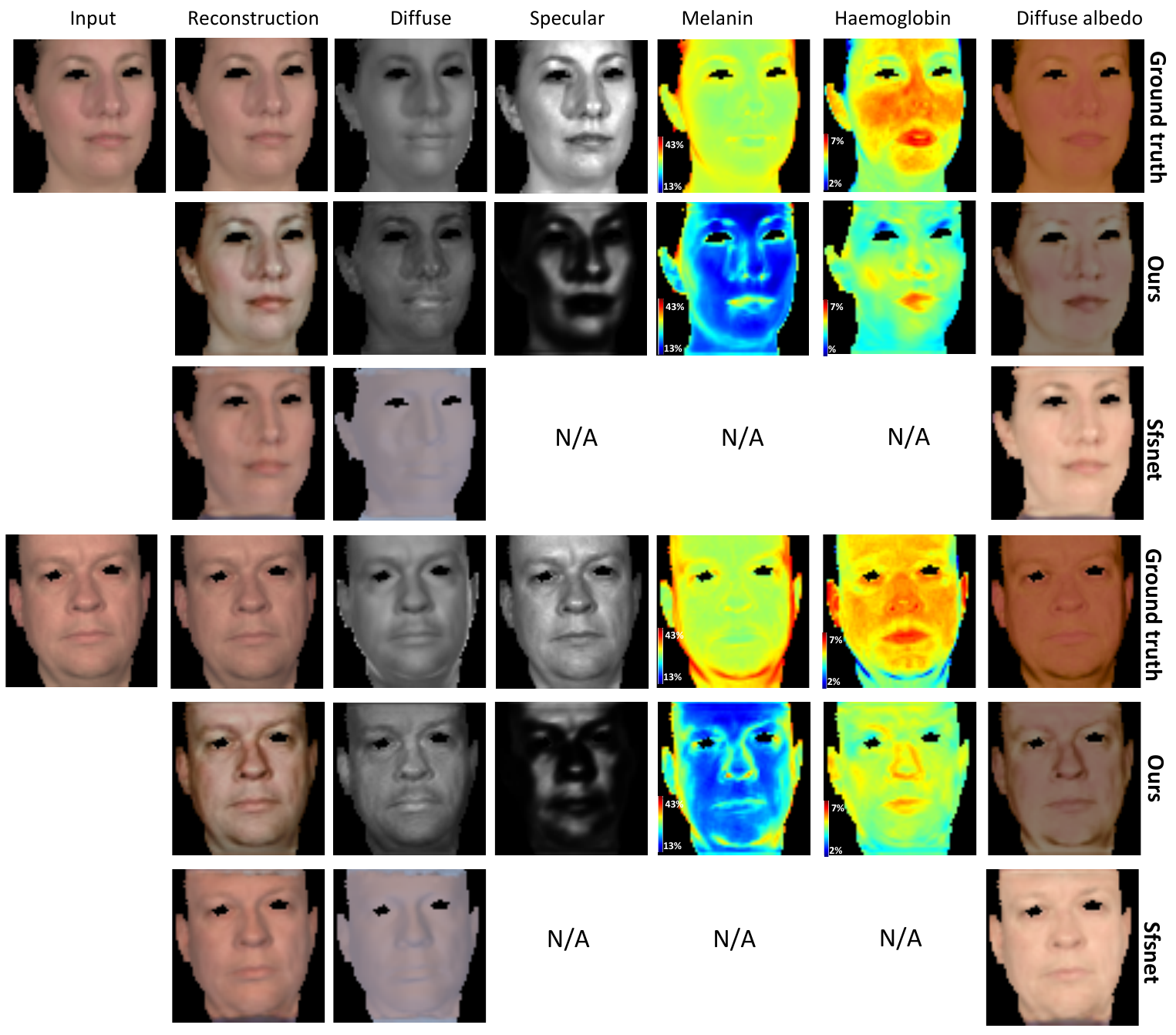}
\caption{Qualitative evaluation using multispectral faces dataset, from left to right: sRGB render of multispectral image, the reconstructed sRGB, diffuse shading, specular shading, melanin map, haemoglobin map and the diffuse albedo. For each example, the top row is result from \cite{alotaibi2019decomposing}, middle row is our results and the bottom row is the SfSnet results \cite{SfSNet}.}
\label{fig:comparsion}
\end{figure}

BioFaceNet is the first work attempt to decompose real images into biophysical maps and diffuse and specular shading. Moreover, there is no ground truth available for this task since no existing device or method can estimate these quantities from real images. For this reason, we propose a new benchmark based on pseudo ground truth computed from multispectral images but give our network access only to RGB images rendered from the multispectral data. We use the decomposition method proposed by Alotaibi and Smith \cite{alotaibi2019decomposing} and apply it to 25 multispectral face images from the ISET database \cite{ImageVal}. This provides pseudo ground truth for the four maps. We then render the multispectral images to RGB using D65 illumination and the mean camera sensitivity and provide this image as input to our CNN. We measure the RMSE error of each map against the pseudo ground truth. We compare against the diffuse shading and albedo obtained by a state-of-the-art method \cite{SfSNet}. In Fig.~\ref{fig:comparsion} we show qualitative results and in Table \ref{quantitative1} we show quantitative results. Our approach provides better performance than \cite{SfSNet}, though note that our sparse specularity prior is too severe and the albedo appears saturated compared to ground truth.

\section{Conclusions}
We have tackled a highly ambitious task: attempting to decompose a single, uncontrolled image into a biophysical and spectral explanation of the appearance. The main conclusion of our work is that the constraint afforded by restricting reflectance to the space of biophysically plausible skin colours enables a decomposition to be obtained that is qualitatively convincing and quantitatively better than a state-of-the-art inverse rendering method. An obvious extension is to combine this work with methods that estimate 3D face geometry. We currently do not constrain the two shading maps such that they are consistent with an underlying geometry and illumination environment. This additional constraint may improve performance and help disambiguate the task. We would also like to explore whether the intrinsic parameter maps can be used for recognition and whether a recognition loss could be used to help disambiguate the decomposition. Our biophysical colouration model could be made at least partially learnable and adversarial losses could help improve the realism of renderings of the model output (for example by applying transformations to the parameter maps or camera/illumination parameters) while still requiring that the output image is realistic. 

\bibliography{egbib}
\end{document}